# New Vistas to study Bhartṛhari: Cognitive NLP

**Jayashree Aanand Gajjam[†], Diptesh Kanojia[†,♣,⋆], Malhar Kulkarni[†]**
[†]IIT Bombay, India
[♣]IITB-Monash Research Academy, India
[⋆]Monash University, Australia
[†]{jayashree_aanand, diptesh, malhar}@iitb.ac.in

## Abstract

The Sanskrit grammatical tradition which has commenced with *Pāṇini*'s *Aṣṭādhyāyī* mostly as a *Padaśāstra* has culminated as a *Vākyaśāstra*, at the hands of *Bhartṛhari*. The grammarian-philosopher *Bhartṛhari* and his authoritative work *'Vākyapadīya'* have been a matter of study for modern scholars, at least for more than 50 years, since Ashok Aklujkar submitted his Ph.D. dissertation at Harvard University. The notions of a sentence and a word as a meaningful linguistic unit in the language have been a subject matter for the discussion in many works that followed later on. While some scholars have applied philological techniques to critically establish the text of the works of *Bhartṛhari*, some others have devoted themselves to exploring philosophical insights from them. Some others have studied his works from the point of view of modern linguistics, and psychology. Few others have tried to justify the views by logical discussions.

In this paper, we present a fresh view to study *Bhartṛhari*, and his works, especially the *'Vākyapadīya'*. This view is from the field of Natural Language Processing (NLP), more specifically, what is called as Cognitive NLP. We have studied the definitions of a sentence given by *Bhartṛhari* at the beginning of the second chapter of *'Vākyapadīya'*. We have researched one of these definitions by conducting an experiment and following the methodology of silent-reading of Sanskrit paragraphs. We collect the Gaze-behavior data of participants and analyze it to understand the underlying comprehension procedure in the human mind and present our results. We evaluate the statistical significance of our results using T-test, and discuss the caveats of our work. We also present some general remarks on this experiment and usefulness of this method for gaining more insights in the work of *Bhartṛhari*.

## 1 Introduction

Language is an integral part of the human communication process. It is made up of structures. There are sentences, which are made up of words, which in turn are made up of syllables. There has been a lot of discussion about which among these is a minimal meaningful unit in the language. The notions of a sentence and a word have been described in different fields of knowledge such as grammar, linguistics, philosophy, cognitive science etc. Some provide a formal definition of a sentence, while others present the semantic definition. The *Vyākaraṇa*, *Mīmāṃsā* and *Nyāya* schools of thought in Sanskrit literature hold some views about the nature of a sentence. The grammarian-philosopher *Bhartṛhari* enumerated eight definitions of a sentence given by early grammarians and *Mīmāṃsakas* in the second *Kāṇḍa* (Canto) of his authoritative work *'Vākyapadīya'* (fourth century A.D).

The question that how does a human being understand a sentence has been dealt with in the field of psycholinguistics for the last 20 years. Various studies conducted in last decade have addressed this question by using several experimental methods. There are many off-line tasks[1] such as Grammaticality Judgement task, Thematic Role Assignment task etc. which are helpful in examining how the language-users process the complete sentences. In addition to these off-line techniques, psycho-linguists have investigated a number of sophisticated on-line language comprehension methodologies. Some of them are behavioral methods such as Acceptability Judgement, Speed-Accuracy Trade-off, Eye-Movement Behavior, Self-Paced Reading etc. Some are neuro-cognitive methods such as electroencephalogram (EEG)[2], Event-Related brain Potentials (ERPs)[3], functional Magnetic Resonance Imaging (fMRI)[4], Positron Emission Tomography (PET)[5] etc. which study the ongoing or real-time cognitive procedure while a participant performs a task.

This paper addresses one of the eight definitions given by *Bhartṛhari*. The main goal is to study this definition from cognitive point of view i.e. to study the underlying comprehension procedure in the human beings taking this definition as the foundation. It also allows us to find the cases of linguistic behavior of the readers in which this definition holds true. We use Eye Tracker device to collect the Gaze (or Eye) movement data of readers during the procedure of silent reading[6] of Sanskrit paragraphs.

**Gaze Tracking: An Introduction**

Gaze tracking is the process of measuring a gaze point or the movement of the participants' eyes. The device which measures the eye-movements is called as Eye-Tracker. We use an 'SR-Research Eyelink-1000 Plus'[7] which mainly comprises of two PCs (Host PC and Display PC), a camera and an infrared illuminator. It performs the monocular eye-tracking with a sampling rate of 500Hz (one sample/2 millisecond). Host PC is used by the supervisor for navigating through the experiment. Supervisor can set up the camera, perform the eye-calibration process, check and correct the drifts, present the paragraphs to the readers and record the session on the Host PC. Similarly, Display PC is used by the reader for reading the paragraphs and answering the questions. The pupil of the participant is captured by the camera and the eye-movements are captured by the infrared illuminator. These eye-movements are mapped to the data that is presented to the participant on the Display PC with the help of some image processing algorithms.

Eye-Tracker records several eye-movement parameters on the Area of Interest (AOI) such as *Pupil size*, *Fixations* and *Saccades*. An AOI is an area of the display that is *of the concern*, like a word or a sentence or a paragraph, which in our case is a word. A *Fixation* is when the gaze is focused on a particular interest area for 100-500 milliseconds. A *Saccade*[8] is the movement of gaze between two fixations which occurs at an interval of 150-175 milliseconds.[9] Specifically, due to its high sampling rate, Eye-Tracker is also able to capture *Saccadic-Regressions* and similarly *Progressions.* A *Regression* a.k.a *Back-tracking* is a backward-moving saccadic movement in which the reader looks back to something that they had read earlier. On the contrary, a *Progression* is a forward-moving saccadic path.

---

[1] These methodologies are called as 'off-line' techniques because they study the comprehension process after the participant performs the task, most of which are the pen-paper methods focusing on question-answer patter for the analysis.

[2] EEGs measure the electrical activities of the brain while performing a task by applying electrode/s to the scalp.

[3] ERPs provide a very high temporal resolution. The spontaneous electrical activity of the brain as a result of various cognitive processes is measured non-invasively by means of electrodes applied to the scalp (Choudhary, 2011).

[4] fMRIs are BOLD (Blood Oxygen Level Dependent) techniques and used to study both the neurologically healthy adults and people with reading disabilities, mostly the brain-damaged patients. It depends on the fact that different cognitive activities by the human being lead to the difference in the blood-oxygen level in the brain.

[5] PETs are the neuro-imaging techniques which are based on the assumptions that areas of high radioactivity are correlated with the brain activities.

[6] Even though the oral and silent reading represent the same cognitive process, readers decrease processing time on difficult words in silent as compared to oral reading. (Juel and Holmes, 1981). For the current paper, we focus only on the silent-reading methodology of the paragraphs.

[7] More information can be found at the link: `http://www.sr-research.com`

**Gaze-Tracking: Feasibility**

The availability of embedded inexpensive eye-trackers on hand-held devices has come close to reality now. This opens avenues to get eye-tracking data from inexpensive mobile devices from a huge population of online readers non-intrusively, and derive cognitive features. For instance, *Cogisen:* has a patent (ID: EP2833308-A1)[10] on eye-tracking using an inexpensive mobile webcam.

**Gaze-Tracking: Applicability**

Till date, there has been lots of research which have been carried out using eye movement data on various tasks such as reading (texts, poetry, musical notes, numerals), typing, scene perception, face perception, mathematics, physics, analogies, arithmetic problem-solving and various other dynamic situations (driving, basketball foul shooting, golf putting, table tennis, baseball, gymnastics, walking on an uneven terrain, mental rotation, interacting with the computer screens, video game playing etc.) and media communication (Lai et al., 2013) etc. *Reading researcher*s have applied eye-tracking for behavioral studies as surveyed by Rayner (1998). Recently, some researchers have even used this technique to explore learning processes in complex learning contexts such as emergent literacy, multimedia learning, and science problem-solving strategies.

In Section 2, we discuss the related work in the fields of Sanskrit grammatical tradition and cognitive NLP. In the next Section 3, we present our approach which focuses on the experimentation details and we present the analysis and results in Section 4. Section 5 gives the evaluation of our work, which is followed by the Section 6 on discussion. We conclude this paper in Section 7 by suggesting possible future work.

## 2 Related Work

In this section, we discuss the work that has been undertaken on the notions of sentence and sentence-meaning by Indian and Western scholars in subsection 2.1. The studies that have been carried out in the fields of Cognitive NLP are presented in subsection 2.2. We also present a bird's eye view of our research area in the figure at the end of this section.

---

[8]The word 'Saccade' is a French-origin word. It was Luis Émile Javal (French eye specialist and a politician) who named the movement of the eyes as 'Saccades' for the first time in 19th C.

[9]As far as human anatomy is concerned, eyes are never still; there are small movements/tremors of the eyes all the time. They are called as 'Nystagmus' (Rayner, 1998). These eye movements are involuntary and do not reveal anything much about the comprehension in reader's mind. As opposed to this, the movements of the eyes which are deliberate and occur at the interval of 150-175 ms are supposed to be the effect of the reading attentively.These eye-movements are indeed considered as the features for the analysis.

[10]http://www.sencogi.com

## 2.1 Sentence Definitions and Comprehension

Sanskrit grammatical tradition is started with *Pāṇini*'s *'Ashtadhyayi'*. *Pāṇini* in his work doesn't define a sentence explicitly. However, few modern scholars attribute a sentence as the base of the derivational process in *Pāṇini*'s grammar (Kiparsky and Staal, 1969), which view is criticized by few other scholars (Houben, 2008; Joshi and Roodbergen, 2008). According to some scholars, the notion of *Kāraka* (Huet, 2006) or the notion of *Sāmarthya* (Deshpande, 1987; Devasthali, 1974) are *Pāṇini*'s contribution to the syntax. The latter view is opposed by Mahavir (1984). After *Pāṇini*, *Kātyāyana* who wrote *Vārttikas* on the rules of *Aṣṭādhyāyī* gave two definitions of the sentence[11] for the first time, which are said to be formal in their nature and not referring to the meaning content (Matilal, 1966; Pillai, 1971; Laddu, 1980). Deshpande (1987) argued that *Kātyāyana*'s claim that each sentence must have a finite verb relates to the deeper derivational level and not to its surface expressions. Hence, a sentence may or may not contain a finite verb on the surface level and there can be a purely nominal sentence (Bronkhorst, 1990; Coward, 1976; Tiwari, 1997). *Patañjali* in his *'Mahābhāṣya'* discussed the integrity of a sentence in terms of having only one finite verb. According to him, a sentence must have only one finite verb, and also purely nominal sentences may not be considered as complete. The word *'asti'* ('is') should be understood in those sentences (Bronkhorst, 1990). Modern scholars discussed that a sentence having two identical finite verbs[12] doesn't militate against the integrity of a sentence (Pillai, 1971; Jha, 1980; Laddu, 1980; Deshpande, 1987).

**Bhartṛhari**, for the first time, deals with the semantic issues in the second *Kāṇḍa* i.e *Vākyakāṇḍa* of *Vākyapadīya* (VP). We can observe a comprehensive treatment on various theories of sentence and their meanings along with their philosophical discussions. He enumerates eight views on the notion of a sentence which are held by earlier theorists in India. The verse is:

*Ākhyātaśabdaḥ saṅghāto jātiḥ saṅghātavartinī*
*Eko'navayaḥ śabdaḥ kramo buddhyanusaṃhṛtiḥ |*
*Padamādyaṃ pṛthaksarvaṃ padaṃ sākāṅkṣamityapi*
*Vākyaṃ prati matirbhinnā bahudhā nyāyavādinam ||* (VP.II.1-2)

The meaning of these definitions is as follows: (1) *Ākhyātaśabdaḥ*- The verb, (2) *Saṅghātaḥ*- A combination of words, (3) *Jātiḥ saṅghātavartinī*- The universal in the combination of words, (4) *Eko'navayavaḥ śabdaḥ*- An utterance which is one and devoid of parts, (5) *Kramaḥ*- A sequence of words, (6) *Buddhyanusaṃhṛtiḥ*- The single whole meaning principle in the mind, (7) *Padamādyam*- The first word, and (8) *Pṛthak sarvam padam sākāṅkṣam*– Each word having expectancy for one another. These eight views on the sentence are held by earlier grammarians and *Mīmāṃsakas*. They look at the sentence from different angles depending upon the mental dispositions formed due to their discipline in different *Śāstras*.[13]

The definitions *'jātiḥ saṅghātavartinī'*, *'eko'navayavaḥ śabdaḥ'* and *'buddhyanusaṃhṛtiḥ'* can be categorized under *Bhartṛhari*'s theory of *'Sphoṭa'* which believes that a sentence is 'a single undivided utterance' and its meaning is 'an instantaneous flash of insight'. This definition is studied by various modern scholars in their respective works. (Raja, 1968; Pillai, 1971; Coward, 1976; Sriramamurti, 1980; Tiwari, 1997; Loundo, 2015). Some modern scholars have studied the theory of *'Sphoṭa'* in different perspectives. Coward (1973) showed the logical consistency and psychological experience[14] of *'Sphoṭa'* theory, while Houben (1989) compared *Bhartṛhari*'s *Śabda* to Saussure's theory of sign (Houben, 1989).[15] Later on, Akamatsu (1993) tried to look at this theory in the philosophical and historical context of the linguistic theory in India.

---

[11] *'ākhyātaṃ sāvyayakārakaviśeṣaṇaṃ vākyam |'* (P.2.1.1 Vt.9) (Meaning: 'A sentence is chiefly the action-word, accompanied by the particle, nominal words, and adjectives.') and *'ekatiṅ vākyam |'* (P.2.1.1 Vt.10) (Meaning: 'A sentence is that [cluster of words] containing a finite verb [as an element]').

[12] The definition *'ekatiṅ vākyam'* is explained by *Patañjali* by giving the illustration of *'brūhi brūhi'*, which indicates that a verb repeated is to be regarded as the same. *Kaiyyaṭa*, the commentator on the *Mahābhāṣya*, also takes the term *'eka'* as 'identical'.

[13] *'Avikalpe'pi vākyārthe vikalpā bhāvanāśrayāḥ' |* (VP II.116)

In contrast with the theory of *'Sphoṭa'*, *Mīmāṃsaka*s hold the view that a syllable has a reality of its own and the word is a sum-total of the syllables and the sentence is only words added together. The remaining definitions such as *'ākhyātaśabdaḥ'*, *'saṅghātaḥ'*, *'kramaḥ'*, *'padamādyam'* and *'pṛthak sarvam padam sākāṅkṣam'* are categorized under this view. Modern Indian scholars (Bhide, 1980; Jha, 1980; Iyer, 1969; Gangopadhyay, 1993; Sriramamurti, 1980; Choudhary, 2011) have discussed the compositionality of a sentence in their respective works. This view is also studied by Western psycho-linguists such as Sanford and Sturt (2002), and criticized by Pagin (2009) who asserts that it is not enough to understand the meanings of the words to understand the meaning of the whole sentence. Studies by Foss and Hakes (1978), Davison (1984), Glucksberg and Danks (2013) and Levy et al. (2012) proved that the sequence is the important parameter in understanding the English sentence. Similar studies by McEuen (1946) and Davison (1984) have shown that people usually tend to skip the first word in the sentence unless it is semantically loaded.

We study the very first definition i.e. **'ākhyātaśabdaḥ'** which states that a single word *'ākhyāta'* ('The Verb') is the sentence. The explanation of this definition as given by *Bhartṛhari* himself in VP.II.326 suggests that if a mere verb denotes the definite means of the action (i.e. the agent and accessory) in the sentence then that verb should also be looked upon as a sentence.[16] In the introduction to the *Ambākartrī* commentary on the VP by Pt. Raghunatha Sarma, he discusses this view by giving examples such as *'pidhehi'*. He mentions that when someone utters the mere verb i.e. *'pidhehi'* ('Close' [imperative]), it also necessarily conveys the *'karma'* of the action which is *'dvāram'* ('the door'), in which case, the mere verb *'pidhehi'* can be considered as a complete sentence[17] (Sarma, 1980). This view is emphasized by later modern scholars by saying that if a linguistic string is to be considered as a sentence, it should have the expectancy on the level of the semantics and not just on the word-level (Pillai, 1971; Laddu, 1980). As stated by the commentator *Puṇyarāja*, this definition believes that the meaning of a sentence is of the nature of an action[18], which means **the meaning of the finite verb becomes the chief qualificand in the cognition that is generated** and other words in the sentence confirm that understanding of a particular action[19] (Pillai, 1971; Huet, 2006). Moreover, as said in the commentary, this definition does not deny the status of the sentence of the linguistic string which contains other words besides the verb. But it emphasizes the fact that, sometimes a single verb can also convey the complete meaning, hence can be looked upon as a sentence.[20] Depending upon these views established by the commentary, we can explain the word **'ākhyātaśabdaḥ'** in both ways viz. the compound *'ākhyātaśabdaḥ'* is analyzed either

---

[14]Coward argues that, according to traditional Indian *Yoga*, the *'Sphoṭa'* view of language is practically possible. It is both logically consistent and psychologically realizable.

[15]Houben suggested that in both the works, a purely mental signifier plays an important role.

[16] "*ākhyātaśade niyataṃ sādhanaṃ yatra gamyate |*
*tadapyekaṃ samāptārthaṃ vākyamityabhidhīyate ||*" (VP.II.326)

[17]*pidhehīti... atra dvāramiti karmākṣepāt paripūrṇārthatve 'dvāraṃ pidhehi' iti vākyam bhavatyeva |*

[18]*'kriyā vākyārhtaḥ' |*

[19] "*Kriyā kriyāntarādbhinnā niyatādhārasādhanā |*
*Prakrāntā pratipattṛṇāṃ bhedaḥ sambodhahetavaḥ ||*" (VP.II.414)

[20]*'tatrākhyātaśabdo vākyamti vādinām ākhyātaśabda eva vākyamiti nābhiprāyaḥ... kintu kvacid ākhyātaśabdo'pi vākyam, yatra kārakaśabdaprayogaṃ vinā kevlākhyātaśabdaprayoge'pi vākyārthāvagatiḥ...'* (*Ambākartrī* on VP.II.1-2)

as *'ākhyātaḥ eva śabdaḥ'* (i.e. *Karmadhāraya Samāsa*- 'The verb' [itself can also be considered as a sentence.]) or as *'ākhyātaḥ śabdaḥ yasmin tat'* (i.e. *Bahuvrīhi Samāsa*- 'the linguistic string consisting the verb' [is a sentence.]),[21] both of which are qualified as 'a sentence'. However, one cannot decide whether this definition leaves out purely nominal sentences when it comes to assign the status of the sentence.[22]

Some earlier work on this view in the field of Psycholinguistics such as McEuen (1946) prove that in the English language, the sentence cognition takes place even if the verb is unavailable. The same view is put forward later by Choudhary (2011). He showed that in verb-final languages such as Hindi, comprehenders do not wait for the verb in case they have not been reached to it yet but they process the sentence incrementally. The study by Osterhout et al. (1994) showed that the verb has the complement-taking properties. Hence, it is the major element in the procedure of sentence-comprehension.

Considering these studies as the motivation, we test the definition of the verb by using an experimental method i.e. by using readers' **Eye Movement Behavior** on the data which contains sentences with verbs, without verbs (purely nominal sentences) and which lack the agents. We are aware that there might be some shortcomings with this definition. There can be the cases or situations in which this definition doesn't hold true or holds true partially.[23] However, *the current research would result in eliciting the cases in which it does.* Hence, we carry out an experiment to find out the situation in which this definition is valid and also provide statistical evidence for the same.

## 2.2 Cognitive NLP

It is very clear from the vast number of studies that Eye Movement behavior can be used to infer cognitive processes (Groner, 1985; Rayner, 1998; Starr and Rayner, 2001). *'The eye is said to be the window into the brain'* as quoted by Majaranta and Bulling (2014). Rayner (1998) has mentioned in his work that the reading experiments have been carried out in different languages such as English, French, Dutch, Hebrew, German (Clematide and Klenne, 2013), Finnish, Japanese and Chinese etc. There are few studies on Indian languages such as Hindi (Choudhary, 2011; Husain et al., 2014; Ambati and Indurkhya, 2009; Joshi et al., 2013) and on Telugu (Ambati and Indurkhya, 2009). The writing style is mainly from left to right except for Hebrew (right to left). Khan et al. (2017) studied the eye movement behavior on Urdu numerals which is written bidirectionally. The orthography has been both horizontal and vertical (Japanese and Chinese). These works have been taken place at various levels of language such as typographical, orthographical, phonological (Miellet and Sparrow, 2004), lexical (Husain et al., 2014), syntactic (Fodor et al., 1974), semantic, discourse, stylistic factors, anaphora and coreference (Rayner, 1998). Few studies were conducted on fast readers versus poor readers, children versus adults versus elderly adults, multilinguals versus monolinguals (De Groot, 2011), normal readers versus people with reading disabilities such as dyslexia, aphasia (Levy et al., 2012), brain damages or clinical disability (Rayner, 1998), schizophrenia, Parkinson's disease (Caplan and Futter, 1986) or oculomotor diseases. Various methodologies were followed such as eye contingent display change, moving window technique, moving mask technique, boundary paradigm, naming task, Rapid Serial Visual Presentation (RSVP) versus Self-paced reading, reading silently versus reading aloud etc.

---

[21]We, in this paper, have studied the latter view, and have presented the sentences having verbs along with the other words as the stimuli to the participants. Studying the first view would require presenting the only-verb sentences which would have led to the loss of context when it comes to the written language cognition. Hence, in stead of presenting only-verb sentences, we have modified the data in such a way that results would help us find out whether the verbs can carry all the semantic load of the sentence in absence of their complements.

[22]We also tried to present these kind of sentences, to study if the purely nominal sentences are apprehended as effortlessly as the sentences having verbs, or whether these sentences amount to the excessive cognitive load in the readers which makes the readers to search for the verb for the better apprehension of the sentence.

[23]Such as in poetry, some concern is also to be given to the sequence (*'kramaḥ'*) of the words. While learning new language, every word including first word (*'padamādyaṃ'*) seems to play the major role etc.

The experiments that took place on reading have been used mainly to understand the levels underlying the comprehension procedure. Apart from that, a study for word sense disambiguation for the Hindi Language was performed by Joshi et al. (2013) where they discuss the cognitive load and difficulty in disambiguating verbs amongst other part-of-speech categories. They also present a brief analysis of disambiguating words based on different ontological categories. Martinez-Gómez and Aizawa (2013) use Bayesian learning to quantify reading difficulty using readers' eye-gaze patterns. Mishra et al. (2013) propose a framework to predict difficulty in translation using translator's eye-gaze patterns. Similarly, Joshi et al. (2014) introduce a system for measuring the difficulties perceived by humans in understanding the sentiment expressed in texts. From a computational perspective Mishra et al. (2016a) predict the readers' sarcasm understandability, detect the sarcasm in the text (Mishra et al., 2017b) and analyze the sentiment in a given sentence (Mishra et al., 2016b) by using various features obtained from eye-tracking.

Eye tracking has been used extensively for Natural Language Processing (NLP) applications in the field of Computer Science, apart from the immense amount of studies done in the field of psycholinguistics. Mishra et al. (2017c) model the complexity of a scan path, and propose the quantification of lexical and syntactic complexity. They also perform sentiment and sarcasm classification (Mishra et al., 2017a) using neural networks using eye tracking data via the use of a convolutional neural network (CNN) (LeCun and others, 1998). They refer to the confluence of attempting to solve NLP problems via cognitive psycholinguistics as *Cognitive NLP*.

Our method of analyzing eye-movement patterns in the Sanskrit language is a *first of its kind* and is inspired by these recent advancements.

The *Bird's eye view* of our research area is presented in Figure 1. The highlighted and bold text is our research interest for the current paper.

## 3 Our Approach

We describe our approach to dataset creation in Subsection 3.1, experiment details which includes participant selection in Subsection 3.2, feature description in Subsection 3.3, followed by the methodology of the experiment in Subsection 3.4.

### 3.1 Dataset Creation

We prepare a dataset of twenty documents consisting of either a prose (Total 13) or a poetry (a *subhāṣita*) (Total 7) in the Sanskrit language. Prose documents mainly contain the stories taken from the texts such as *Pañcatantra*, *Vaṃśavṛkṣaḥ* and *Bālanītikathāmālā* while as *Subhāṣita*s are taken from the text *Subhāṣitamañjūṣā*. The stories are comprised of 10-15 lines each, and each *subhāṣita* is 2 - 4 verse long. We create three copies of 20 paragraphs as the experiment demands and manipulate them as follows:

- **Type A:** These are twenty documents which do not contain any changes from the original documents. They are kept as they were.

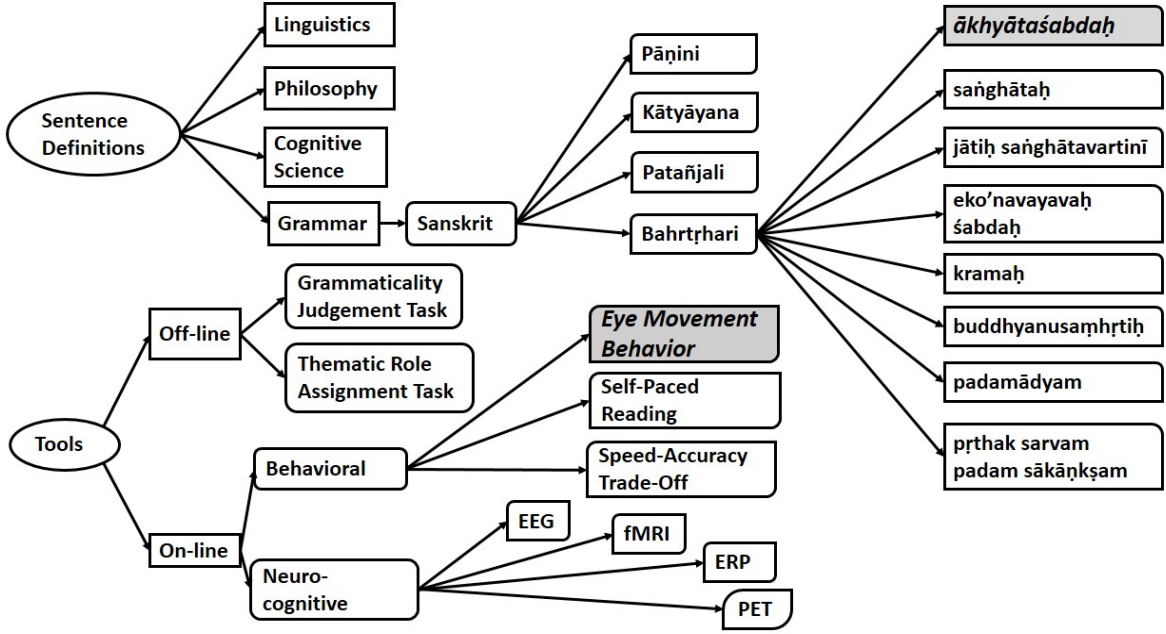

Figure 1: A bird's eye view of our research area

- **Type B:** In this set of documents, we remove the finite and infinite verbs completely which results in a syntactic violation in the respective sentences. These are purely nominal sentences. In poetry, instead of removing the verbs, we replace the verbs with its synonym verb to maintain the format of the poetry. The motivation behind this kind of modification is to test how much does a verb contribute to the comprehension of a sentence, both syntactically and semantically. There are twenty documents of this kind.

- **Type C:** Here, the verbs are kept constant but we drop the *kartā* in the sentences. *kartā* being semantically loaded in the sentence, we choose to drop it as the experiment demands i.e. to investigate whether a mere verb without its agent can denote the complete meaning of the sentence. *Kartās* are not removed from the sentences which did not have finite or infinite verbs in the original document to avoid the possibility of insufficient information while reading. This kind of modification will throw some light on the view that verb itself can be considered as a sentence. In Type C of poetry, the stimulus is degraded by replacing the original finite verbs by distant-meaning finite verbs by retaining the same grammatical category. Even though these verbs bear the syntactic integrity of the sentence, they tend to be semantically incompatible with the other words in the linguistic string. This incompatibility leads to the semantic inhibition while processing it, which in turn allows the reader to reconstruct the meaning of the sentence all over again. There are twenty documents of this kind.

The paragraphs do not contain text which readers might find difficult to comprehend. We normalize the text to avoid issues with vocabulary. We control the orthographical, typographical and lexical variables that might affect the outcome of the experiment. We maintain a constant orthography throughout the dataset. The passages are shown in *Devanāgarī* script and the writing style is from left to right. We keep the font size large, customize the line spacing to optimum and adjust the brightness of the screen for the comfort of the participant. We ensure that there is no lexical complexity in the prose. We minimize it by splitting the *sandhis* (total 70), separating the compound words with the hyphens (total 51) and also by adding commas in appropriate places for the easier reading. The verses are not subject to this kind of modification. This forms our original document. Sentences in the original dataset vary in their nature with respect to the verbs. There are 7 purely nominal sentences, 33 sentences with no finite verb but the *kṛdantas* and 70 sentences having at least one finite verb in them. There are no single-sentence paragraphs which eliminate the possibility of insufficient contextual information while reading. In poetry, there are 26 finite verbs in total, each verse having 3 to 4 finite verbs in it. Two linguists validate our dataset with 100% agreement that the documents are not incomprehensible. This forms the ground truth for our experiment.

We have counterbalanced the final datasets to remove the bias of the paragraphs. All the three types of documents (i.e. Type A, B, and C) are shuffled in such a way that no reader gets to read both types of the same paragraph. Twenty of such shuffled paragraphs make one final dataset. There are three final datasets: Datasets 1, 2 and 3. Out of the twenty participants, seven participants are presented with *Dataset 1*, six participants with *Dataset 2* and remaining seven participants with *Dataset 3*. We formulated two multiple-choice questions on each paragraph. The first question of which is one and the same for all paragraphs which help us get the reader's viewpoint about the meaningfulness of the paragraph concerned. The second question is based on the gist of that paragraph which works as a comprehension test for the readers, which also ensures that people have read attentively and eliminates the cases of mindless reading. The answers given by the participants on both questions are used by us to decide the inter-annotator agreement and the accuracy rate.

### 3.2 Experiment Details

We chose twenty participants[24] with a background in Sanskrit[25] who were tested individually. They have been learning Sanskrit for minimum 2 years to maximum more than 10 years. The participants are neurologically healthy adults who belong to the age group of minimum 22 to maximum 38 years. They are well-acquainted with the Sanskrit language, however, they were not aware of the purpose of the experiment beforehand. Participants were also naïve about the modifications made to the datasets. All of the participants can understand, read and speak multiple languages. While most of the participants are native speakers of Marathi; few of them have Kannada, Telugu, and Hindi as their native language. They had a normal vision or corrected vision and reported no reading or other language-related disorders.

Before the experiment, all the participants were given a set of instructions mentioning the procedure of the experiment, description of the apparatus, annotation input method, necessity of the head-movement minimization during the experiment and need of the mindful reading etc. They were instructed to take short breaks whenever necessary to avoid the fatigue over the time and also to maintain the concentration during reading. They were given two sample documents before the experiment so that they get to know the working of the experimentation

---

[24] The number of participants may seem less. The primary reason for less number of participants is due to the restriction that we needed our readers to know Sanskrit language. Secondly, we chose the readers with normal or corrected vision since the readers who use bi-focal eyeglasses would pose a minor possibility of erroneous eye-movement data. Moreover, some other human-related aspects such as very dark or very light irises, downward pointing eyelashes, naturally droopy eyelids, the headrest not fitting the person's head or even the incorrigible head motions amount to the calibration fails and errors while reading. We aim to increase the number of participants in future experiments.

[25] We chose to investigate this definition on the Sanskrit data instead of the native languages of the participants. Firstly because, it would be more faithful to study the definition on the language that had been the lingua franca at the time it was elaborated. Secondly, Sanskrit as the second language formed a base for all the participants belonging to the different variety of native languages. Nonetheless, it would be interesting to examine the same definition on the native speakers as mentioned in the Section 7 (future work) and carry out the contrastive study for having better conception of the definition.

process. They were rewarded financially for their efforts.

### 3.3 Feature Description

The eye-tracking device records the activity of the participant's eye on the screen and records various features through gaze data. We do not use all the feature values provided by the device for our analysis, but only the ones which can provide us with the prominence of a word (interest-area) and in turn, show us the importance of words which belong to the same category. These are features which are calculated based on the gaze behavior of the participant, and we use for our analysis:

1. **Fixation-based features** -
   Studies have shown that attentional movements and fixations are obligatorily coupled. More fixations on a word are because of incomplete lexical processes. More cognitive load will lead to more time spent on the respective word. There are some variables that affect the time spent on the word such as word frequency, word predictability, number of meanings of a word or word familiarity etc. (Rayner, 1998). We consider Fixation duration, Total fixation, Fixation Count for the analysis. These are motivated by Mishra et al. (2016a)

   (a) Fixation Duration (or First Fixation Duration)-
   First fixations are fixations occurring during the first pass reading. Intuitively, an increased first fixation duration is associated with more time spent on the words, which accounts for lexical complexity.

   (b) Total Fixation Duration (or Gaze Duration)-
   This is a sum of all fixation durations on the interest areas. Sometimes, when there is syntactic ambiguity, a reader re-reads the already read part of the text in order to disambiguate the text. Total fixations duration accounts for sum of all such fixation durations occurring during the overall reading span.

   (c) Fixation Count-
   This is the number of fixations on the interest area. If the reader reads fast, the first fixation duration may not be high even if the lexical complexity is more. But the number of fixations may increase on the text. So, fixation count may help capture lexical complexity in such cases.

2. **Regression-based feature** -
   Regressions are very common in complicated sentences and many regressions are due to comprehension failures. Short saccade to the left is done to read efficiently. Short within-word saccades show that a reader is processing the currently fixated word. Longer regression (back the line) occur because the reader did not understand the text. Syntactic ambiguity (such as Garden Path sentences etc.), syntactic violation (missing words, replaced words) and syntactic unpredictability leads to shorter saccades and longer regressions. We consider the feature Regression Count i.e. a total number of gaze regressions around the area of interest.

3. **Skip Count** -
   Our brain doesn't read every letter by itself. While reading people keep on jumping to next word. Predictable target word is more likely to be skipped than an unpredictable one. We take Skip count as a feature to calculate the results. Skip count means whether an interest-area was skipped or not fixated on while reading. This is calculated as number of words skipped divided by total word count. Intuitively, higher skip count should correspond to lesser semantic processing requirement (assuming that skipping is not done intentionally).

Two factors have a big impact on skipping: word length and contextual constraint. Short words are much more likely to be skipped than long words. Second, words that are highly constrained by the prior context are much more likely to be skipped than those that are not predictable. Word frequency also has an effect on word skipping, but the effect is smaller than that of predictability.

4. **Run Count** -
   Run count is the number of times an interest-area was read.

5. **Dwell Time-based feature** -
   Dwell time and Dwell Time percentage i.e. the amount of time spent on an interest-area, and the percentage of time spent on it given the total number of words.

### 3.4 Methodology

As described above in Section 3.1, we modified the documents in order to test the syntactic and semantic prominence of a verb in both prose and poetry. Such instances of modification of the data may cause a syntactic violation, semantic inhibition and leads to insufficient information to comprehend the document, at the surface level of the language. It enforces the reader to re-analyze the text. The time taken to analyze a document depends on the context (Ivanko and Pexman, 2003). While analyzing the text, the human brain would start processing the text in a sequential manner, with the aim of comprehending the literal meaning. When such an *incongruity* is perceived, the brain may initiate a re-analysis to reason out such disparity (Kutas and Hillyard, 1980). *As information during reading is passed to the brain through eyes, incongruity may affect the way eye-gaze moves through the text. Hence, distinctive eye-movement patterns may be observed in the case of the successful finding of a verb, in contrast to an unsuccessful attempt.* This hypothesis forms the crux of our analysis and we aim to prove this by creating and analyzing an eye-movement database for sentence comprehension.

## 4 Analysis & Results

As stated above, we collect gaze data from 20 participants and use it for our analysis. We try to verify the first sentence definition given by *Bhartṛhari*. With our work, we find that **the verb** is the chief contributor to the sentence-semantics and enjoys more attention than other words in the process of sentence comprehension. To study how does a reader uses a verb in constructing the meaning of a linguistic string, we analyze the time one spends on the particular verb (dwell-time percentage), the number of times one backtracks (regression out count) or skips (skip count) the verb, the number of times the verb is read through (run count) and fixated upon (fixation count). We analyze these features on the verbs vs. non-verbs in Datasets 1, 2 and 3 and present the results in the Figures 2 (dwell-time percentage), 3 (regression count) and 4 (skip count) in the form of graphs.

The analysis of dwell-time percentage, regression count and skip count proves our point that verbs are prominent element while constructing the sentence meaning. It can be clearly seen that **verbs are spent more time on, regressed about more and skipped a lesser number of times than non-verbs.** All the participants except a few correlate with our hypothesis. We observe that in Figure 2, Participant 5 (P5) has spent less time on the verbs but we also observe, as shown in Table 1, that P5 lacks in agreement compared to the other annotators. Participants 11 (P11), 12 (P12) and 18 (P18) do not lack in agreement, still, they do not read verbs as much as the other consistent participants and hence are clearly outliers. Even though these four participants have not fixated on the verb for more time, the number of times they regressed around verbs is significantly higher as shown in the Figure 3. Figure 4 shows that verbs are unanimously skipped for lesser number of times than non-verbs, hence it is proved that a reader cannot afford to skip verbs while constructing the sentence meaning.

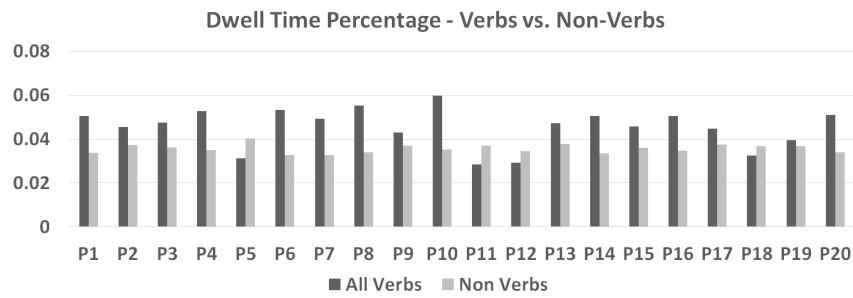

Figure 2: A Comparison of Dwell-Time Percentage on Verbs and Non-Verbs for all Datasets, and all participants

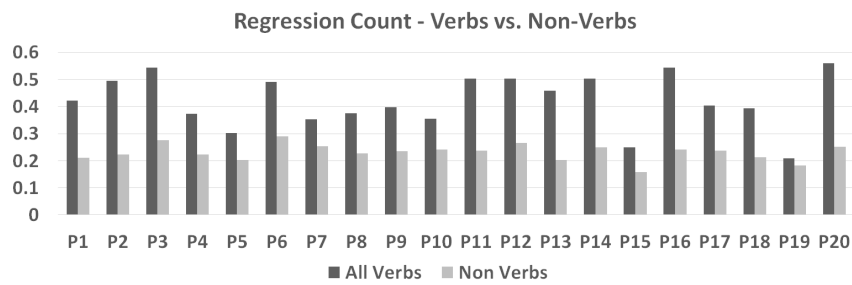

Figure 3: A Comparison of Regression Count on Verbs and Non-Verbs for all Datasets, and all participants

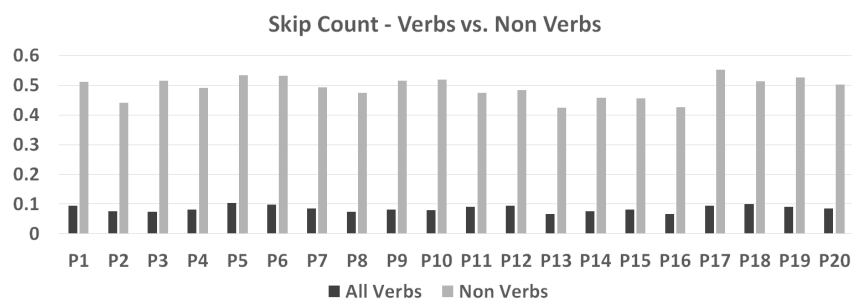

Figure 4: A Comparison of Skip Count on Verbs and Non-Verbs for all Datasets, and all participants

We also strengthen this view by analyzing the Type A vs. Type B vs. Type C documents and also consider the answers provided by the readers in the Section 6.

## 5   Evaluation

We perform the evaluation of our work and calculate inter-annotator agreement (IAA) for each participant with all the others, on the same dataset. We perform this for both the questions posed to the participants, separately. We also evaluate the answers provided by the participants to ensure that none of them were performing an inattentive reading of the documents. We show our evaluation in Tables 1, 2, and 3 for *Dataset 1, 2 and 3* respectively. Overall, the agreement of our participants ranges between **0.45** (Moderate Agreement) to **0.95** (Almost perfect Agreement) for Question 1. For Question 2, the agreement ranges from **0.5** (Moderate Agreement) to **0.95** (Almost perfect Agreement). **The Accuracy (Acc), as shown in the tables, ranges from 0.6 to 1**, which means that our participants were substantially accurate and were attentive during the experiment. The inter-annotator agreement points our the tentative outliers and helps us analyze the results of our experiment. We find that both inter-annotator agreement and accuracy of our experiment are substantial.

We also perform statistical significance tests based on the standard t-test formulation assuming unequal variances for both variables, for all participants and display the *p*-values in Tables 4, 5, 6 for *Datasets 1, 2, and 3* respectively. For these datasets, we compare Verbs with all the other words for the features Regression Count (RC) and Skip Count (SC). We find out that a number of regressions performed by a user around verbs are much more than around other words. For these features, we also show the difference between the means of verbs and non-verbs ($M_D$), and the *p*-value (P). Our T-Test parameters were variable values, the hypothesized mean difference was set to zero, and the expected cut-off for the T-Test is 0.05. Our evaluations show that these values are statistically significant for most of the participants.

**Inter-annotator agreement (IAA) and Accuracy (Acc) Scores**

|  | Q1 | Q2 | |
|---|---|---|---|
|  | IAA | IAA | Acc |
| **P1** | 0.7 | **0.5** | **0.6** |
| **P2** | 0.8 | 0.9 | 0.95 |
| **P3** | 0.8 | 0.9 | 0.9 |
| **P4** | **0.95** | **0.95** | 0.95 |
| **P5** | **0.45** | 0.85 | 0.9 |
| **P6** | 0.9 | 0.55 | **0.6** |
| **P7** | 0.85 | 0.7 | 0.8 |

Table 1: Dataset 1

|  | Q1 | Q2 | |
|---|---|---|---|
|  | IAA | IAA | Acc |
| **P8** | 0.85 | 0.9 | 0.95 |
| **P9** | 0.75 | 0.6 | 0.75 |
| **P10** | 0.75 | 0.8 | 1 |
| **P11** | 0.65 | 0.75 | 0.85 |
| **P12** | 0.7 | 0.8 | 0.85 |
| **P13** | 0.85 | **0.95** | 1 |

Table 2: Dataset 2

|  | Q1 | Q2 | |
|---|---|---|---|
|  | IAA | IAA | Acc |
| **P14** | 0.8 | 0.8 | 0.75 |
| **P15** | 0.65 | 0.65 | 0.75 |
| **P16** | 0.85 | 0.9 | 0.95 |
| **P17** | 0.9 | 0.8 | 0.7 |
| **P18** | 0.75 | 0.85 | 0.85 |
| **P19** | 0.5 | 0.9 | 0.9 |
| **P20** | 0.8 | 0.7 | 0.8 |

Table 3: Dataset 3

**Mean Difference and p-values from T-Test for Regression Count (RC) and Skip Count (SC)**

|  | RC | | SC | |
|---|---|---|---|---|
|  | $M_D$ | P | $M_D$ | P |
| **P1** | 0.159 | 0.000 | 0.061 | 0.038 |
| **P2** | 0.234 | 0.000 | 0.078 | 0.012 |
| **P3** | 0.250 | 0.000 | 0.180 | 0.000 |
| **P4** | 0.126 | 0.001 | 0.112 | 0.001 |
| **P5** | 0.062 | 0.050 | 0.029 | 0.194 |
| **P6** | 0.183 | 0.001 | 0.064 | 0.029 |
| **P7** | 0.091 | 0.029 | 0.089 | 0.005 |

Table 4: Dataset 1

|  | ROC | | SC | |
|---|---|---|---|---|
|  | $M_D$ | P | $M_D$ | P |
| **P8** | 0.141 | 0.001 | 0.129 | 0.000 |
| **P9** | 0.147 | 0.001 | 0.134 | 0.000 |
| **P10** | 0.112 | 0.005 | 0.143 | 0.000 |
| **P11** | 0.194 | 0.000 | 0.025 | 0.237 |
| **P12** | 0.163 | 0.003 | 0.012 | 0.364 |
| **P13** | 0.211 | 0.000 | 0.106 | 0.001 |

Table 5: Dataset 2

|  | ROC | | SC | |
|---|---|---|---|---|
|  | $M_D$ | P | $M_D$ | P |
| **P14** | 0.188 | 0.000 | 0.058 | 0.053 |
| **P15** | 0.072 | 0.033 | 0.058 | 0.053 |
| **P16** | 0.244 | 0.001 | 0.077 | 0.015 |
| **P17** | 0.129 | 0.003 | 0.055 | 0.059 |
| **P18** | 0.120 | 0.030 | -0.030 | 0.189 |
| **P19** | 0.021 | 0.247 | 0.044 | 0.106 |
| **P20** | 0.253 | 0.002 | 0.059 | 0.049 |

Table 6: Dataset 3

# 6   Discussion

We discussed the core features of our work *i.e.* Dwell-time Percentage, Regression Count, Skip Count, Run Count, and Fixation Count in Section 4. In this section, we would like to further analyze the result of work by exploring the answers provided by our participants. We break down our documents into the categories of *prose* and *poetry*. In Figures 5a and 5b, we show the answer counts of our participants, when they find the documents absolutely non-meaningful, or lacking information *i.e.,* somewhat meaningful. For all participants, over document Types A, B, and C, we find that Type A (Original Data) is marked non-meaningful least number of times.

In case of a *prose* (Figure 5a), Type B documents lack verbs. It can clearly be seen that our participants do not understand the documents most of the times, and mark them either as completely non-meaningful or lacking in information. We do not hint them to look for verbs as psycholinguistic principles do not allow an experiment to be biased in the participants' mind. Non-presence of verbs in Type B documents affects both syntax and the semantics of the documents and it can be seen that purely nominal sentences fail to convey the complete semantics of the sentence. In Type C for prose (Figure 5a), we see that our participants are confused by the removal of *agent-denoting* words, but are still able to grasp the context, and hence their answers do not depict an absolute meaninglessness of the documents. Even though verbs are retained in document type C, the removal of *agent* words leads to insufficient information.

For *poetry* (Figure 5b), Type B documents have the presence of synonymous verbs, and Type C have verbs with very distant meanings and no correlation with the semantics of the original verb present. Hence, Type B documents are marked as lacking in information by our participants many times as compared to Type A documents. They do not mark even one of them as absolutely meaningless as a synonym of a verb is present and they are still able to grasp the context which bears a strong impact on the conclusion we draw. On a similar note, Type C documents which have verbs but with very distant meanings are marked lacking in information most number of times, as a correlation cannot be established between the expected sense of the original verb and the current verb present in the document.

We explore further and manually analyze the saccadic paths of our participants to find out that in document types A, B, and C, the *saccadic-regressions* vary as per our hypothesis. We present a sample in Figures 6a, 6b and 6c. For a randomly chosen single participant, who has above average IAA and good accuracy, we find that the amount of regression on document Type C increases in comparison to Type A since the document lacks a agent in some sentences. But, for Type B, we can observe that the regressions increase further when the verb is completely removed from the document.

As stated before, the definition that we have studied might not be valid in all the cases. Our aim is to find out the cases in which it does. In the conclusion of this research, we can say that, we have found one such case in which *Bhartṛhari*'s definition *Ākhyātaśabdaḥ* is valid and that is: *when the lexical complexity is minimized in the Sanskrit texts*, readers rely on the verbs in order to understand the complete meaning of the sentence, without which the sentence-meaning seems incomplete. Hence, we can conclude that **verbs play *the most* important role in the syntax and semantics of a sentence**, nonetheless, in most of the cases, they demand their complements (i.e. means of action) to represent the complete semantics of a sentence. We can also conclude that the *purely nominal sentences in Sanskrit are less meaningful* than the corresponding original sentences.

Similarly, we would also like to present Figures 7 (Run Count) and 8 (Fixation Count) which further strengthen our discussion. We can see in both the figures that a number of times a verb has been read is always more than the number of time other words have been read.

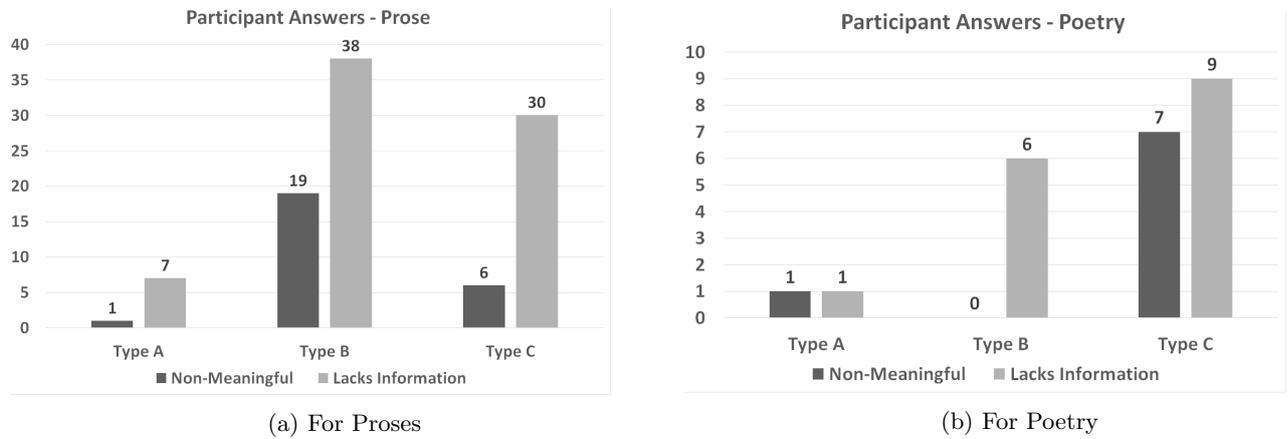

(a) For Proses

(b) For Poetry

Figure 5: Meaninglessness of documents as reported by Participants on different document sets

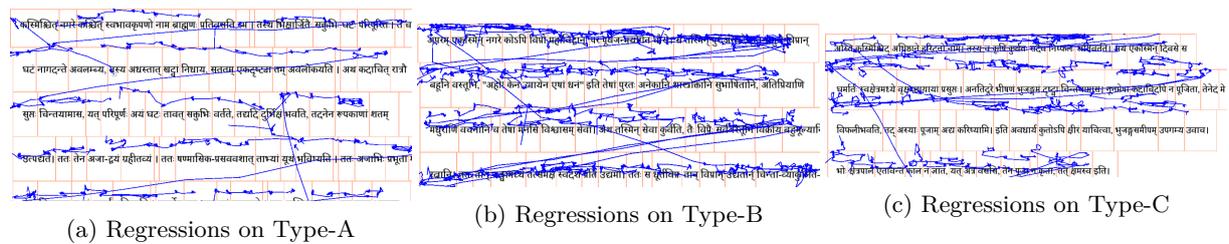

(a) Regressions on Type-A

(b) Regressions on Type-B

(c) Regressions on Type-C

Figure 6: Regression sample from a participant

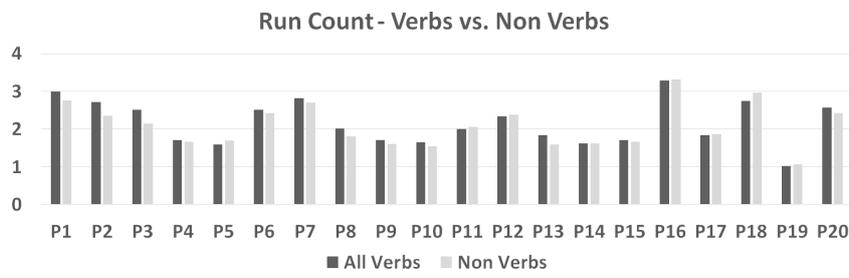

Figure 7: A Comparison of Run Count on Verbs and Non-Verbs for all Datasets, and all participants

**Limitations**

The data selected for our experiment does not vary in its nature. We only use stories in prose, and the poetry is also borrowed from the same text. We would like to clearly state that we know this is a limitation of our work. It will be more insightful to conduct similar experiments on different kinds of texts. For the same experiment on 'verbs', data can also be modified in many other ways. Moreover, a spoken word, when accompanied by gesture and facial expression and when given a special intonation, can convey much more than the written word. This experiment it limited to the written sentences only and it tests the comprehension only from the reader's point of view.

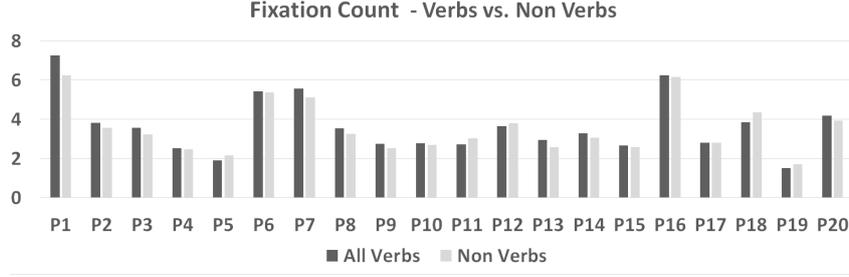

Figure 8: A Comparison of Fixation Count on Verbs and Non-Verbs for all Datasets, and all participants

# 7 Conclusion & Future Work

We present a fresh view to study *Bhartṛhari*'s *'Vākyapadīya'*, especially the definitions given by him on the syntactic and the semantic level. We pick sentence definition one viz. *Ākhyātaśabdaḥ*, that the "verb" can also be considered as a sentence. We discuss his work in brief and perform an experiment to study this definition in cognitive point of view. We employ eye-tracking technique and follow the methodology of silent-reading of Sanskrit paragraphs to perform the above-mentioned experiment in order to have the better understanding of the definition. We aim to extend our work under the purview of Cognitive NLP and use it to resolve computational problems. With our work, we open a new vista for studying sentence definitions in the cognitive point of view by following an investigational technique.

Our results show that humans tend to read verbs more than they read other words and they are deemed most important. We assert that verbs play a prominent role in the syntax and semantics of a sentence, nonetheless, in most of the cases, they demand their complements to represent the complete semantics of a sentence. It is proved that a human being, cognitively, searches for a verb in a sentence, without which the unity of a sentence tends to be incomplete. Purely nominal sentences in the Sanskrit language are less meaningful than the original sentences. We show the statistical significance of our results and evaluate them using the standard T-test formulation. We also discuss the manual analysis of saccadic paths and answer given by our participants to verify our results. We are aware that, the method followed by us is one way of justifying *Bhartṛhari* and there could be other ways which can strengthen the same results.

In future, we aim to conduct more experiments on different kinds of texts in the Sanskrit language which have different sentence-construction styles. For the same experiment, data can also be modified in other ways such as- changing the place of the verb in the sentence, removing the sentence boundary markers, replacing the conjunctions, negatives, discourse markers etc. We also aim to verify other sentence definitions using eye-tracking. We would like to employ other tools such as EEG and work in multi-lingual settings to further delve deeper into the cognition of a human mind so that we can understand the definition in better perspective. We would also like to study the comprehension among the native speakers vs. bilingual so that we can study whether the definitions by *Bhartṛhari* are generic in nature. We hope to gain more insights into the field of Cognitive NLP with the help of our work.

## Acknowledgements

We thank our senior colleague Dr. Abhijit Mishra who provided insights and expertise that greatly assisted this research. We are grateful to a research scholar Vasudev Aital for his assistance in the data-checking process and all the participants for being the part of this research. We would also like to extend out gratitude to the reviewers and the editors for their valuable comments on an earlier version of the manuscript which indeed shaped it better, although any errors are our own.